%% file: PaperForReview.tex
\documentclass[10pt,twocolumn,letterpaper]{article}

\usepackage{cvpr}      

\usepackage{graphicx}
\usepackage{amsmath}
\usepackage{amssymb}
\usepackage[accsupp]{axessibility}
\usepackage{booktabs}
\usepackage{multirow}
\usepackage{subfiles}
\usepackage{color, soul}
\usepackage{bm}
\usepackage{bbm}
\usepackage[pagebackref,breaklinks,colorlinks]{hyperref}

\usepackage[capitalize]{cleveref}
\crefname{section}{Sec.}{Secs.}
\Crefname{section}{Section}{Sections}
\Crefname{table}{Table}{Tables}
\crefname{table}{Tab.}{Tabs.}

\def\cvprPaperID{10893} 
\def\confName{CVPR}
\def\confYear{2023}

\input{sections/defs}

\begin{document}

\title{Logical Implications for Visual Question Answering Consistency}

\author{
Sergio Tascon-Morales \quad Pablo Márquez-Neila \quad Raphael Sznitman\\
University of Bern\\
{\tt\small \{sergio.tasconmorales, pablo.marquez, raphael.sznitman\}@unibe.ch}
}
\maketitle


\input{sections/00_Abstract}
\input{sections/01_Introduction}
\input{sections/02_RelatedWork}

\input{sections/03_Method}
\input{sections/05_Experiments}
\input{sections/055_Results} 
\input{sections/06_Conclusion}
\input{sections/07_Acknowledgements}

{\small
\bibliographystyle{ieee_fullname}
\bibliography{egbib}
}

\end{document}

%% file: sections/defs.tex
\newif\ifdraft
\drafttrue

\definecolor{orange}{rgb}{1,0.5,0}
\definecolor{pink}{rgb}{0.98, 0.38, 0.5}
\definecolor{darkgreen}{rgb}{0.055, 0.490, 0.016} 

\ifdraft
 \usepackage[normalem]{ulem}
 \newcommand{\RS}[1]{{\color{red}{\bf RS: #1}}}
 
 \newcommand{\PMN}[1]{{\color{orange}{\bf PMN: #1}}}
 
 \newcommand{\STM}[1]{{\color{blue}{\bf STM: #1}}}
 
\else
 \newcommand{\sout}[1]{}
 \newcommand{\RS}[1]{{\color{red}{}}}
 
 \newcommand{\PMN}[1]{{\color{red}{}}}
 
 \newcommand{\STM}[1]{{\color{red}{}}}
 
\fi

\newcommand{\x}{\mathbf{x}}

\newcommand{\q}{\mathbf{q}}

\DeclareMathOperator*{\argmax}{arg\,max}

%% file: sections/00_Abstract.tex
\begin{abstract}
Despite considerable recent progress in Visual Question Answering (VQA) models, inconsistent or contradictory answers continue to cast doubt on their true reasoning capabilities. However, most proposed methods use indirect strategies or strong assumptions on pairs of questions and answers to enforce model consistency. Instead, we propose a novel strategy intended to improve model performance by directly reducing logical inconsistencies. To do this, we introduce a new consistency loss term that can be used by a wide range of the VQA models and which relies on knowing the logical relation between pairs of questions and answers. While such information is typically not available in VQA datasets, we propose to infer these logical relations using a dedicated language model and use these in our proposed consistency loss function. We conduct extensive experiments on the VQA Introspect and DME datasets and show that our method brings improvements to state-of-the-art VQA models, while being robust across different architectures and settings.  
\end{abstract}

%% file: sections/01_Introduction.tex
\section{Introduction}
\label{sec:intro}

Visual Questioning Answering (VQA) models have drawn recent interest in the computer vision community as they allow text queries to question image content. This has given way to a number of novel applications in the space of model reasoning~\cite{wang2015explicit,cadene2019murel,wu2021multi,jing2022maintaining}, medical diagnosis~\cite{nguyen2019overcoming,vu2020question,gupta2021hierarchical,zhan2020medical} and counter factual learning~\cite{agarwal2020towards,chen2020counterfactual,abbasnejad2020counterfactual}. With ability to combine language and image information in a common model, it is unsurprising to see a growing use of VQA methods.

Despite this recent progress however, a number of important challenges remain when making VQAs more proficient. For one, it remains extremely challenging to build VQA datasets that are void of bias. Yet this is critical to ensure subsequent models are not learning spurious correlations or shortcuts~\cite{teney2020unshuffling}. This is particularly daunting in applications where domain knowledge plays an important role (\eg,  medicine~\cite{he2020pathvqa,lau2018dataset,do2021multiple}). Alternatively, ensuring that responses of a VQA are coherent, or {\it consistent}, is paramount as well. That is, VQA models that answer differently about similar content in a given image, imply inconsistencies in how the model interprets the inputs. A number of recent methods have attempted to address this using logic-based approaches~\cite{gokhale2020vqa}, rephrashing~\cite{shah2019cycle}, question generation~\cite{ribeiro2019red,ray2019sunny, goel2021iq} and regularizing using consistency constraints~\cite{tascon2022consistency}. In this work, we follow this line of research and look to yield more reliable VQA models.

\begin{figure}[!t]
\centering
\includegraphics[width=0.4\textwidth]{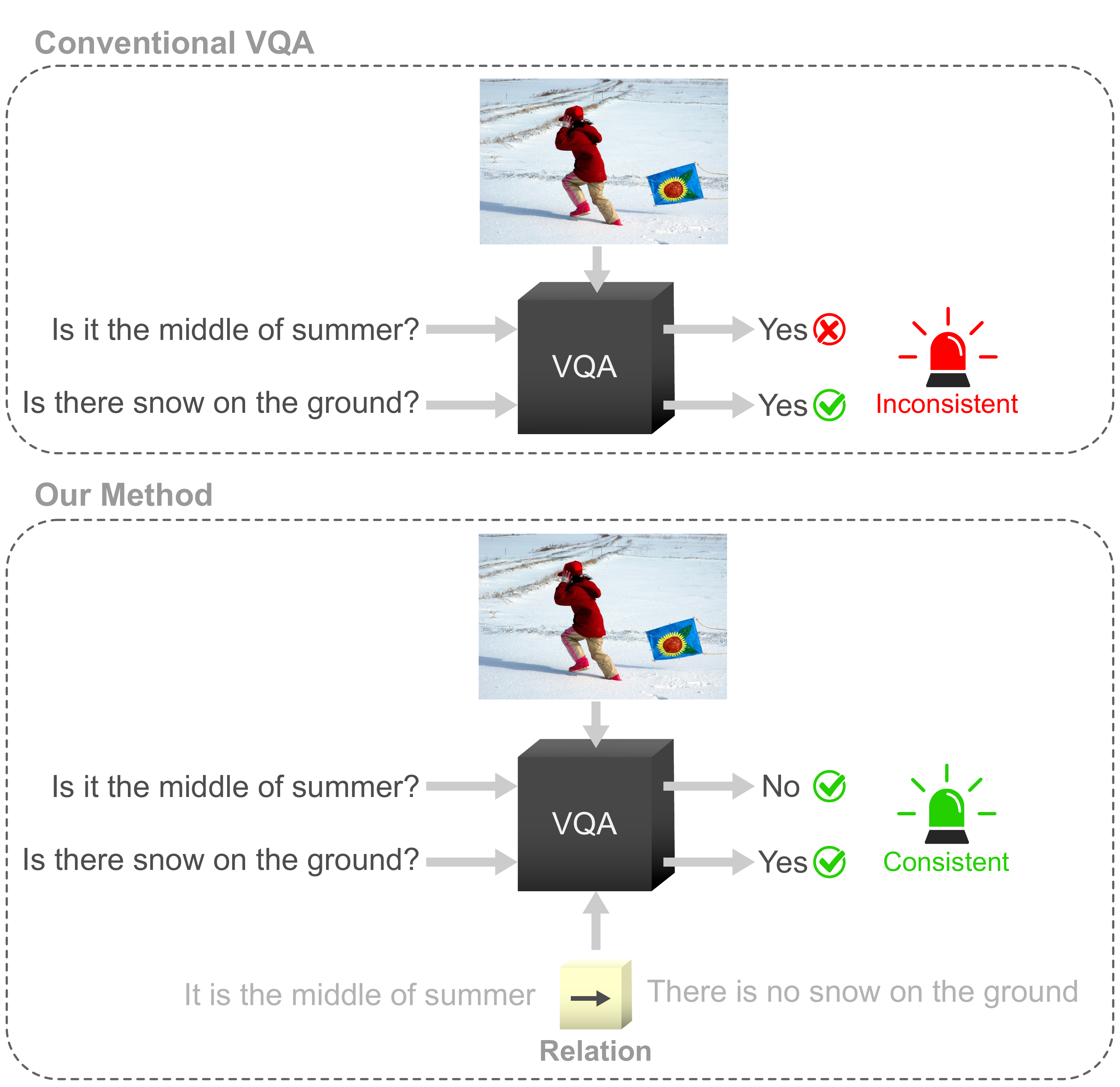}
\caption{Top: Conventional VQA models tend to produce inconsistent answers as a consequence of not considering the relations between question and answer pairs. Bottom: Our method learns the logical relation between question and answers pairs to improve consistency.
} 
\label{fig:image_intro}
\end{figure}

Specifically, we wish to ensure that VQA models are consistent in their ability to answer questions about images. This implies that if one poses multiple questions about the same image, then the model's answers should not contradict themselves. For instance, if one question about the image in Fig.~\ref{fig:image_intro} asks ``Is there snow on the ground?", then the answer inferred should be consistent with that of the question ``Is it the middle of summer?" As noted in~\cite{selvaraju2020squinting}, such question pairs involve reasoning and perception, and consequentially lead the authors to define inconsistency when the reasoning and perception questions are answered correctly and incorrectly, respectively. Along this line,~\cite{tascon2022consistency} use a similar definition of inconsistency to regularize a VQA model meant to answer medical diagnosis questions that are hierarchical in nature. What is critical in both cases however is that the consistency of the VQA model depends explicitly on its answers, as well as the question and true answer. This hinges on the assumption that perception questions are sufficient to answer reasoning question. Yet, for any question pair, this may not be the case. As such, the current definition of consistency (or inconsistency) has been highly limited and does not truly reflect how VQAs should behave. 

To address the need to have VQA models that are self-consistent, we propose a novel training strategy that relies on logical relations. To do so, we re-frame question-answer (QA) pairs as propositions and consider the relational construct between pairs of propositions. This construct allows us to properly categorise pairs of propositions in terms of their logical relations. From this, we introduce a novel loss function that explicitly leverages the logical relations between pairs of questions and answers in order to enforce that VQA models be self-consistent.
Unfortunately however, datasets typically do not contain relational information about pairs of QA and collecting this would be extremely laborious and difficult to achieve. To overcome this, we propose to train a dedicated language model capable of inferring logical relations between propositions.
By doing so, we show in our experiments that not only are we able to effectively infer logical relations from propositions, but that these can be explicitly used in our loss function to train VQA models that improve state-of-the-art methods via consistency. We show this over two different VQA datasets, against different consistency methods and with different VQA model architectures. Our code and data are available at \url{https://github.com/sergiotasconmorales/imp_vqa}.

%% file: sections/02_RelatedWork.tex
\section{Related work}
\label{sec:relatedwork}

Since its initial presentation in Antol \etal~\cite{antol2015vqa}, VQA has thoroughly advanced. Initial developments focused on multimodal fusion modules, which combine visual and text embeddings~\cite{nam2017dual,cadene2019murel}. From basic concatenation and summation~\cite{antol2015vqa}, to more complex fusion mechanisms that benefit from projecting the embeddings to different spaces, numerous approaches have been proposed~\cite{fukui2016multimodal,kim2016hadamard,ben2017mutan}. The addition of attention mechanisms~\cite{kim2018bilinear, nam2017dual,cadene2019murel} and subsequently transformer architectures~\cite{vaswani2017attention} has also contributed to the creation of transformer-based vision-language models, such as LXMERT, which have shown state-of-the-art performances~\cite{tan2019lxmert}. 

More recently, methods have proposed to improve other aspects of VQA, including avoiding shortcut learning and biases~\cite{dancette2021beyond,han2021greedy}, improving 3D spatial reasoning~\cite{banerjee2021weakly}, Out-Of-Distribution (OOD) generalization~\cite{cao2021linguistically,teney2020unshuffling}, improving transformer-based vision-language models~\cite{yang2021auto,zhou2021trar}, external knowledge integration~\cite{ding2022mukea,gao2022transform} and model evaluation with visual and/or textual perturbations~\cite{gupta2022swapmix,walmer2022dual}. With the awareness of bias in VQA training data some works have also addressed building better datasets (\eg, v2.0~\cite{goyal2017making}, VQA-CP~\cite{agrawal2018don}, CLEVR~\cite{johnson2017clevr} and GCP~\cite{hudson2019gqa}).

Furthermore, these developments have now given rise to VQA methods in specific domains. For instance, the VizWiz challenge~\cite{gurari2018vizwiz,gurari2019vizwiz,chen2022grounding} aims at creating VQA models that can help visually impaired persons with routine daily tasks, while there is a growing number of medical VQA works with direct medicine applications~\cite{nguyen2019overcoming,gupta2021hierarchical,vu2020question,zhan2020medical}. 

\paragraph{Consistency in VQA}
Consistency in VQA can be defined as the ability of a model to produce answers that are not contradictory. This is, given a pair of questions about an image, the answers predicted by a VQA model should not be contrary (\eg answering ``Yes" to ``Is it the middle of summer?" and ``Winter" to ``What season is it?"). Due to its significance in reasoning, consistency in VQA has become a focus of study in recent years~\cite{ribeiro2019red,shah2019cycle,gokhale2020vqa,selvaraju2020squinting,jing2022maintaining}. Some of the first approaches for consistency enhancement focused on creating re-phrasings of questions, either by dataset design or at training time~\cite{shah2019cycle}. Along this line, entailed questions were proposed~\cite{ribeiro2019red,gokhale2020vqa}, such that a question generation module was integrated into a VQA model~\cite{ray2019sunny,goel2021iq}, used as a benchmarking method to evaluate consistency~\cite{yuan2021perception} or as a rule-based data-augmentation technique~\cite{ribeiro2019red}. Other approaches tried to shape the embedding space by imposing constraints in the learned representations~\cite{teney2019incorporating} and by imposing similarities between the attention maps of pairs of questions~\cite{selvaraju2020squinting}. Another work~\cite{tascon2022consistency} assumed entailment relations between pairs of questions to regularize training. A more recent approach attempts to improve consistency by using graph neural networks to simulate a dialog in the learning process~\cite{jing2022maintaining}. 

While these approaches show benefits in some cases, they typically only consider that a subset of logical relationships exists between pairs of question-answers or assume that a single relation holds for all QA pairs. Though true in the case of re-phrasings, other question generation approaches cannot guarantee that the produced questions preserve unique relations or that grammatical structure remains valid. Consequently, these methods often rely on metrics that either over or under estimate consistency by relying on these assumptions. In the present work, we propose a strategy to alleviate these limitation by considering all logical relations between pairs of questions and answers. 

\paragraph{Entailment prediction} Natural Language Inference (NLI), or Recognizing Textual Entailment (RTE), is the task of predicting how two input sentences (namely \textit{premise} and \textit{hypothesis}) are related, according to three pre-established categories: entailment, contradiction and neutrality~\cite{maccartney2008modeling}. For example, if the premise is ``A soccer game with multiple males playing" and the hypothesis is ``Some men are playing a sport," then the predicted relation should be an entailment, because the hypothesis logically follows from the premise. Several benchmarking datasets (\eg, SNLI~\cite{young2014image}, MultiNLI~\cite{williams2017broad}, SuperGLUE~\cite{wang2019superglue}, WIKI-FACTCHECK~\cite{sathe2020automated} and ANLI~\cite{nie2019adversarial}) have contributed to the adaption of general-purpose transformer-based models like BERT~\cite{devlin2018bert}, RoBERTa~\cite{liu2019roberta} and DeBERTa~\cite{he2020deberta} for this task. In this work, we will leverage these recent developments to build a model capable of inferring relations between propositions.

%% file: sections/03_Method.tex
\section{Method}
\label{sec:method}
Given an image $\x \in \mathcal{I}$, a question $\q \in \mathcal{Q}$ about the image and a set $\mathcal{A}=\{a_1,\ldots, a _K\}$ of possible answers to choose from, a VQA model is expected to infer the answer $\hat{a} \in \mathcal{A}$ that matches the true answer $a^*$. This can be formulated as, 
\begin{equation}
    \hat{a} = \argmax_{a \in \mathcal{A}} p(a | \x,\q; \theta),
    \label{eq:vqa}
\end{equation}
where $\theta$ represents the parameters of the VQA model.


In this context, we observe that two QA pairs $(\q _i, a_i)$ and~$(\q_j, a_j)$ for the same image~$\x$ can have different kinds of logical relations. In the simplest case, the two pairs may be unrelated, as with the pairs (``Is it nighttime?'', ``Yes'') and (``Is there a bench in the image?'', ``No''). Knowing that one of the pairs is true gives no information about the truth value of the other. 

On the other hand, two pairs may be related by a logical implication, as in the pairs (``Is the horse brown?", ``No") and (``What is the color of the horse?", ``White"). Knowing that the second pair is true implies that the first pair must be true as well. Conversely, if the first pair is false (\emph{the horse is brown}), it  implies that the second pair must also be false. In this case, the first pair is a necessary condition for the second one or, equivalently, the second pair is a sufficient condition for the first one. 

Finally, it can be that two QA pairs are related by a double logical implication, as with the pairs (``Is this a vegetarian pizza?'', ``Yes'') and (``Does the pizza have meat on it?'', ``No''). The veracity of the former implies the veracity of the latter, but the veracity of the latter also implies the veracity of the former. In this case, each pair is simultaneously a necessary and sufficient condition for the other pair, and both pairs are then equivalent. 

Note that the logical implication existing between two QA pairs is an intrinsic property of the QA pairs, and does not depend on the correctness of the predictions coming from a VQA model. If a VQA model considers a sufficient condition true and a necessary condition false, it is incurring an \emph{inconsistency} regardless of the correctness of its predictions.

Since logical implications are the basis of reasoning, we propose to explicitly use them when training a VQA model to reduce its inconsistent predictions.
Unfortunately, doing so requires overcoming two important challenges: (1)~a strategy is needed to train VQA models with logical relations that leverage consistency in a purposeful manner. Until now, no such approach has been proposed; (2)~VQA datasets do not typically contain logical relations between pairs of QA. Acquiring these manually would, however, be both time-consuming and difficult.

We address these challenges in this work by formalizing the idea of consistency and treating QA pairs as logical propositions from which relations can comprehensively be defined. Using this formalism, we first propose a strategy to solve (1) and train a VQA model more effectively using logical relations and the consistency they provide~(\cref{subsec:loss}). We then show in~\cref{subsec:relation_prediction} how we infer relations between pairs of propositions, whereby allowing standard VQA datasets sets to be augmented with logical relations.



\subsection{Consistency formulation}
\label{subsec:consistency_framework}

We begin by observing that QA pairs~$(\q, a)$ can be considered and treated as logical propositions.
For instance, the QA (``Is it winter?", ``Yes") can be converted to ``It is winter," which is a logical proposition that can be evaluated as \emph{true} or \emph{false} (\ie,~its \emph{truth value}). Doing so allows us to use a broad definition of consistency, namely one that establishes that two propositions are inconsistent if both cannot be true at the same time~\cite{bradley1979possible}. In the context of this work, we assume the truth value of a proposition~$(\q, a)$ is determined by an agent (either a human annotator or the VQA model) after observing the information contained in an image~$\x$. 

Let $\mathcal{D} = \mathcal{I}\times \mathcal{Q} \times \mathcal{A}$ be a VQA dataset that contains triplets $(\x^{(n)}, \q_i^{(n)}, a_i^{(n)})$, where $\x^{(n)}$ is the $n$-th image and $(\q_i^{(n)}, a_i^{(n)})$ is the $i$-th question-answer pair about $\x^{(n)}$. In the following, we omit the index~$n$ for succinctness. For a given image $\x$, we can consider a pair of related question-answers as $(\q_i, a_i)$ and $(\q_j, a_j)$ as a pair of propositions. 
Following propositional logic notation, if both propositions are related in such a way that $(\q_i, a_i)$~is a sufficient condition for the necessary condition $(\q_j, a_j)$, we write that $(\q_i, a_i)\rightarrow (\q_j, a_j)$. For convenience, this arrow notation can be adapted to indicate different orderings between the necessary and sufficient conditions:
\begin{itemize}
    \item $(\q_i, a_i) \leftarrow (\q_j, a_j)$ if the proposition $(\q_i,a_i)$ is a necessary condition for $(\q_j,a_j)$. 
    \item $(\q_i, a_i) \leftrightarrow (\q_j, a_j)$ if the propositions $(\q_i,a_i)$ and $(\q_j,a_j)$ are equivalent, \ie, both are simultaneously necessary and sufficient. Note that this is just notational convenience for the double implication $(\q_i, a_i) \rightarrow (\q_j, a_j) \wedge (\q_j, a_j) \rightarrow (\q_i, a_i)$, and in the following derivations the double arrow will be always considered as two independent arrows.
    \item Finally, we will write $(\q_i, a_i) - (\q_j, a_j)$ if the propositions $(\q_i,a_i)$ and $(\q_j,a_j)$ are not related.
\end{itemize}

If a VQA model is asked questions $\q_i$ and $\q_j$ about an image $\x$ and there exists a relation $(\q_i, a_i) \rightarrow (\q_j, a_j)$, the answers of the model will be inconsistent whenever it provides answers $\hat{a}_i = a_i$ and $\hat{a}_j \neq a_j$ (\ie, the model evaluates the first proposition as true and the second proposition as false). More generally, for a pair of necessary and sufficient conditions, the agent will be inconsistent if it evaluates the necessary condition as false and the sufficient condition as true~\cite{bradley1979possible}.
In what follows, we exploit these ideas to quantify model inconsistencies in our experiments and to develop a new loss function that encourages logically consistent VQA models.


\subsection{Logical implication consistency loss}
\label{subsec:loss}



The core aim of our method is to encourage the VQA model to avoid inconsistent answers. When training, assume that the model receives an image~$\x$ from~$\mathcal{D}$ and two associated propositions~$(\q_1, a_1)$ and~$(\q_2, a_2)$ that are related by a logical implication~$(\q_1, a_1) \rightarrow (\q_2, a_2)$. We define,
\begin{equation}
    \pi_i=\pi \left((\q_i, a_i), \x \right) = p(a_i\mid \x, \q_i, \theta),
\end{equation}
as the probability assigned by the VQA model that the proposition~$(\q, a)$ is true for the image~$\x$. The model has a high probability of incurring an inconsistency if it simultaneously gives a high probability~$\pi_1$ to the sufficient condition and a low probability~$\pi_2$ to the necessary condition.

\begin{figure}
    \centering
    \includegraphics[width=0.75\linewidth]{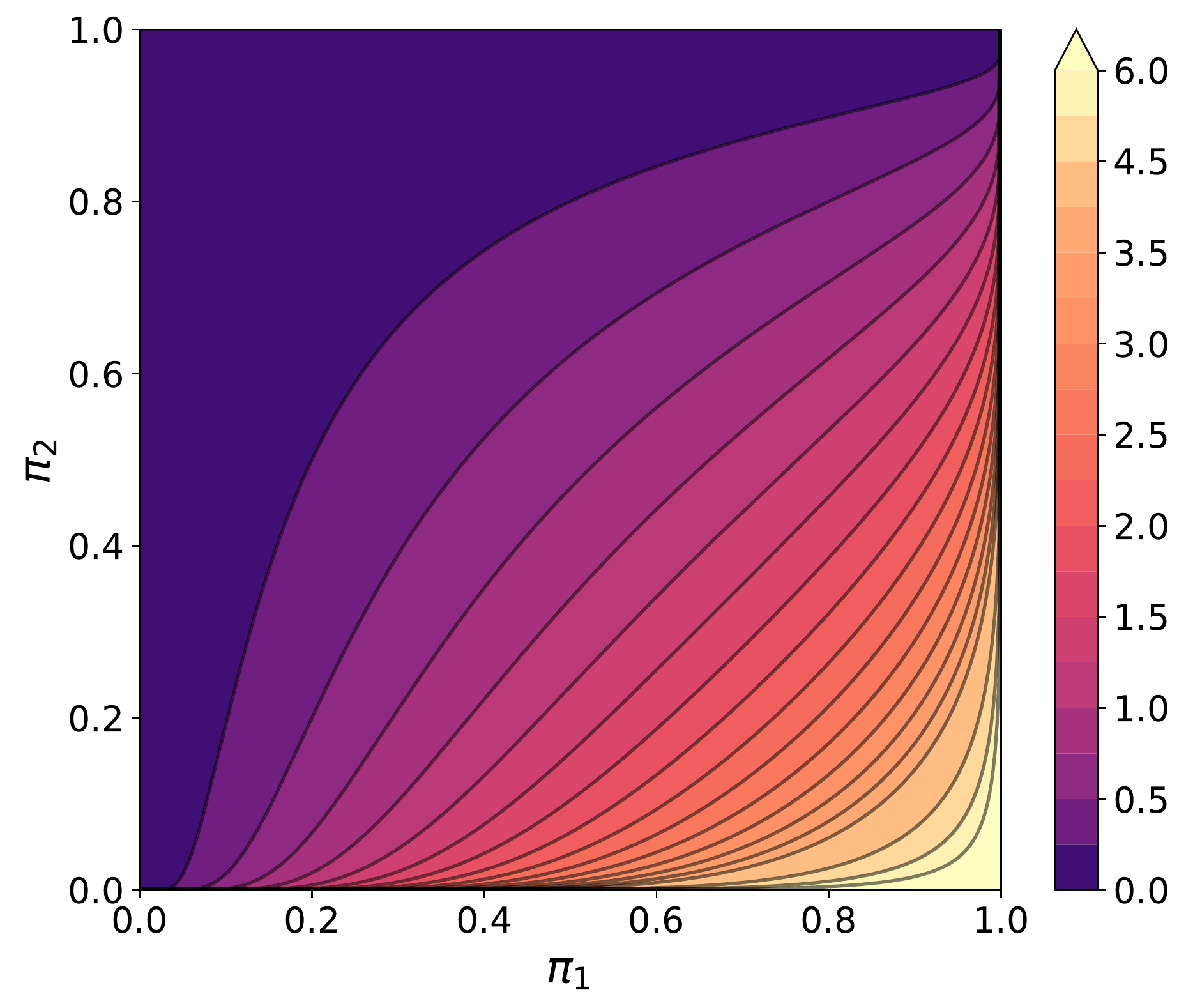}
    \caption{Consistency loss~$\mathcal{L}_\textrm{cons}$ as a function of the estimated probabilities for the sufficient,~$\pi_1$, and necessary,~$\pi_2$, conditions. Note that the loss diverges to~$\infty$ when $\pi_1=1, \pi_2<1$ and when~$\pi_1>0, \pi_2=0$.}
    \label{fig:loss}
\end{figure}

We thus define our consistency loss as a function,
\begin{equation}
\begin{split}
    \mathcal{L}_\textrm{cons}(\x, (\q_1, a_1), (\q_2, a_2)) = - (1-\pi_2) \log(1-\pi_1) \\ - \pi_1 \log(\pi_2),
\end{split}
\end{equation}
that takes an image and a pair of sufficient and necessary propositions, and penalizes predictions with a high probability of inconsistency. As illustrated in \cref{fig:loss}, $\mathcal{L}_\textrm{cons}$~is designed to produce maximum penalties when $\pi_1=1$ and~$\pi_2<1$ (\ie,~when the sufficient condition is absolutely certain but the necessary condition is not), and when $\pi_2=0$ and~$\pi_1>0$ (\ie,~when the necessary condition can never be true but the sufficient condition can be true). At the same time, $\mathcal{L}_\textrm{cons}$ produces minimum penalties when either $\pi_1=0$ or~$\pi_2=1$, as no inconsistency is possible when the sufficient condition is false or when the necessary condition is true. Interestingly, despite its resemblance, $\mathcal{L}_\textrm{cons}$ is not a cross-entropy, as it is not an expectation over a probability distribution.


Our final loss is then a linear combination of the consistency loss and the cross-entropy loss $\mathcal{L}_{\textrm{VQA}}$ typically used to train VQA models. Training with this loss then optimizes, 
\begin{equation}
    \min_\theta \mathbb{E}_{\mathcal{D}}[\mathcal{L}_{\textrm{VQA}}] +
    \lambda\mathbb{E}_{\substack{((\x_i, \q_i, a_i), (\x_j, \q_j, a_j))\sim\mathcal{D}^2 \\ \x_i=\x_j, (\q_i, a_i)\rightarrow (\q_j, a_j)}}[\mathcal{L}_{\textrm{cons}}],
\end{equation}
where the first expectation is taken over the elements of the training set~$\mathcal{D}$ and the second expectation is taken over all pairs of necessary and sufficient propositions from~$\mathcal{D}$ defined for the same image.
In practice, we follow the sampling procedure described in~\cite{selvaraju2020squinting,tascon2022consistency}, where mini-batches contain pairs of related questions. The hyperparameter~$\lambda$ controls the relative strength between the VQA loss and the consistency term. 


\subsection{Inferring logical implications}
\label{subsec:relation_prediction}
\begin{figure}[!t]
\centering
\includegraphics[width=0.475\textwidth]{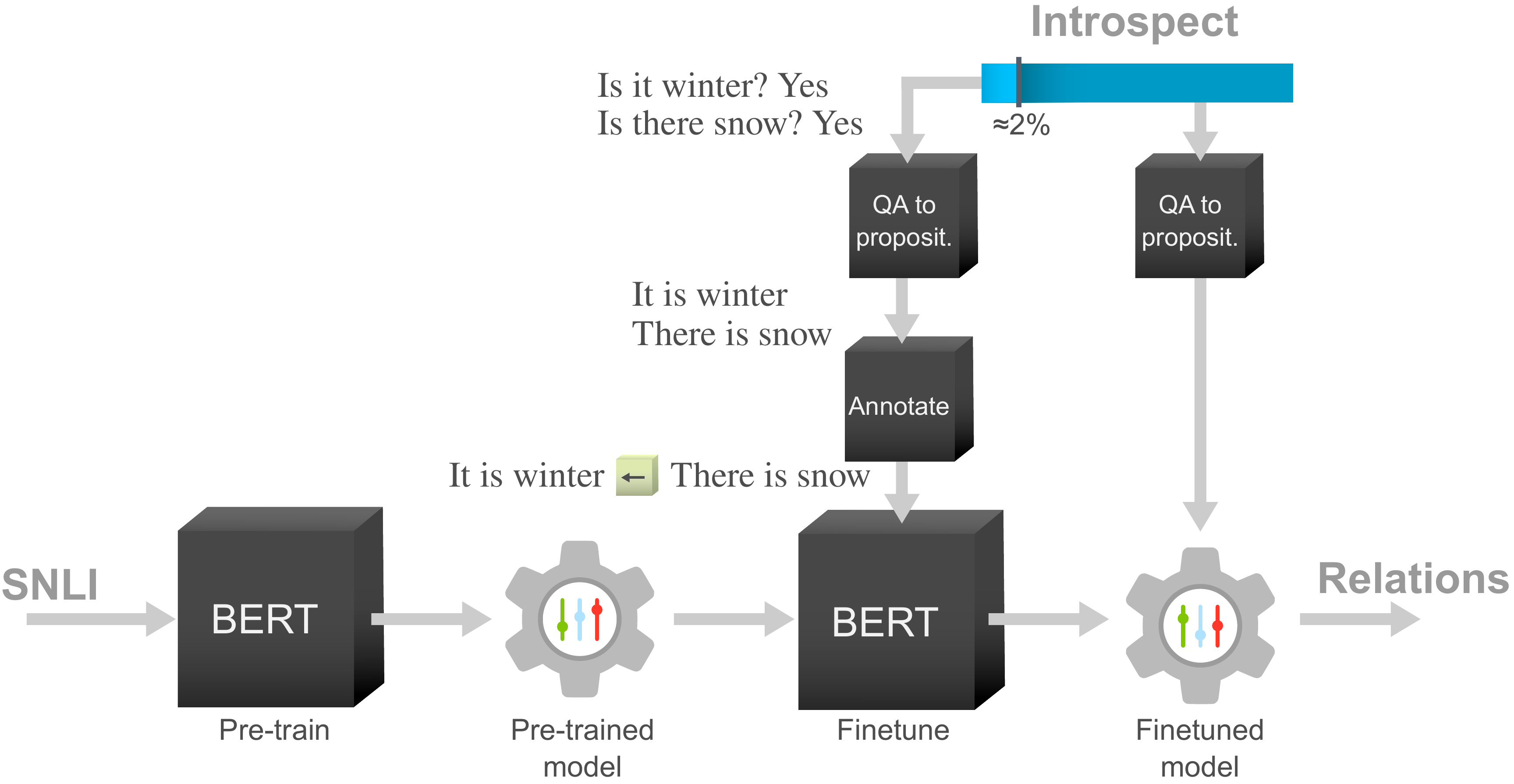}
    \caption{LI-MOD: Approach to predict logical relations between pairs of propositions. A BERT-based NLP model is first pre-trained on the SNLI dataset~\cite{young2014image} to solve a Natural Language Inference task and subsequently fine-tuned fine-tuned with annotated pairs from a subset of Introspect dataset~\cite{selvaraju2020squinting}. The resulting model is used to predict the relations of the remaining part of the dataset.}
\label{fig:relation_prediction}
\end{figure}

By and large, VQA datasets do not include annotations with logical relations between question-answers pairs, which makes training a VQA with $\mathcal{L}_{\textrm{cons}}$ infeasible. To overcome this, we propose to train a language model to predict logical implications directly and use these predictions instead. We achieve this in two phases illustrated in~\cref{fig:relation_prediction} and refer to our approach as the Logical-Implication model (LI-MOD).

First, we pre-train BERT~\cite{devlin2018bert} on the task of Natural Language Inference, using the SNLI dataset~\cite{young2014image}, which consists of pairs of sentences with annotations of entailment, contradiction or neutrality. In this task, given two sentences, a language model must predict one of the mentioned categories. While these categories do not exactly match the logical implication relevant to our objective, it can be derived from the entailment category. To this end, given two propositions $(\q_i,a_i)$ and $(\q_j,a_j)$, we evaluate them using the finetuned NLI model in the order $(\q_i,a_i),(\q_j,a_j)$, and then repeat the evaluation by inverting the order, to evaluate possible equivalences or inverted relations. If the relation is predicted as neutral in both passes, the pair is considered to be unrelated.

Then, we finetune the NLI model on a sub-set of annotated pairs from the VQA dataset Introspect~\cite{selvaraju2020squinting}. In practice, we use a subset of binary QA pairs that were manually annotated with logical implications. Even though the relation need not be limited to binary questions (\ie, yes/no questions), we chose to do so because the relation annotation is simpler than for open-ended questions. Since BERT expects sentences and not QA pairs, these were first converted into propositions using Parts Of Speech (POS) tagging~\cite{petrov2011universal} and used simple rules that apply to binary questions (\eg,~to convert ``Is it winter?," ``Yes" we invert the first two words of the question and remove the question mark). 
After finetuning the model, the relations were predicted for the remaining part of the dataset. Further implementation details on this are given in~\cref{subsec:imp_details}.

%% file: sections/05_Experiments.tex
\section{Experiments}
\label{sec:experiments}

We evaluate our proposed consistency loss function on different datasets, using a variety of VQA models and compare our performances to state-of-the-art methods.

\input{sections/04_Datasets}

\subsection{Quantifying consistency}

Given a test set~$\mathcal{T}= \{t_n\}_{n=1}^{|\mathcal{T}|}$, where $t_n=(\x, \q, a)$ is a test sample triplet, we wish to measure the level of consistency of a VQA model. To this end, we define~$G(\mathcal{T}) \subset \mathcal{T}^2$ as the set of all pairs of test samples~$((\x_i, \q_i, a_i), (\x_j, \q_j, a_j))$ for which~$(\q_i, a_i) \rightarrow (\q_j, a_j)$ and~$\x_i=\x_j$. 

We count the inconsistencies produced by a VQA model~$p$ evaluated on~$\mathcal{T}$ as the number of times the model evaluates a sufficient condition as true and a necessary condition as false,
\begin{equation}
    I_p(\mathcal{T}) = \sum_{(t_i, t_j)\in G(\mathcal{T})} \mathbbm{1}[e_p((\q_i, a_i), \x) \wedge \neg e_p((\q_j, a_j), \x)].
\end{equation}
The function~$e_p$ returns the truth value of the proposition~$(\q, a)$ for image~$\x$ evaluated by the VQA model~$p$,
\begin{equation}
    e_p((\q, a), \x) = \mathbbm{1}[\hat{a}=a],
\end{equation}
where $\hat{a}$ is the answer of maximum probability following Eq.~\eqref{eq:vqa}. In other words, $e_p$~returns whether the estimated answer for question~$\q$ matches the answer of the proposition~$a$.
Finally, the consistency ratio~$c$ for model~$p$ on the test set~$\mathcal{T}$ is the proportion of implications in~$G(\mathcal{T})$ that did not lead to an inconsistency,
\begin{equation}
    c_p(\mathcal{T}) = 1 - \dfrac{I_p(\mathcal{T})}{G(\mathcal{T})}.
\end{equation}

%% file: sections/04_Datasets.tex
\subsection{Datasets}
\label{subsec:datasets}

\paragraph{Introspect~\cite{selvaraju2020squinting}:}Contains perception questions (or sub-questions) created by annotators for a subset of reasoning question (or main questions) of the VQA v1.0 and VQA v2.0 datasets~\cite{antol2015vqa,goyal2017making}. It contains 27,441 reasoning questions with 79,905 sub-questions in its training set and 15,448 reasoning questions with 52,573 sub-questions for validation. For images that have the same sub-question repeated multiple times, we remove duplicates in the sub-questions for every image in both the train and validation sets.

\paragraph{DME Dataset~\cite{tascon2022consistency}:}This dataset consists of retinal fundus images, for the task of Diabetic Macular Edema (DME) staging. It contains 9,779 QA pairs for training, 2,380 QA pairs for validation and 1,311 QA pairs for testing. There are three types of questions in the dataset: main, sub, and independent questions. Main questions ask information about the diagnosis (\ie the stage of the disease) and sub-questions ask about presence and location of biomarkers. Sub-questions are further sub-divided into grade questions, questions about the whole images, questions about a region of the eye called macula, and questions about random regions in the image. To deal with questions about image regions, we follow the procedure described in~\cite{tascon2022consistency}, whereby only the relevant region is shown to the model.


\subsection{Baseline methods and base models} We consider 3 different consistency enhancement baseline methods. To ensure fair comparisons, all methods use the same VQA base models and only differ in the consistency method used. These consist in: 
\begin{itemize}
    \item None: Indicating that no consistency preserving method is used with the VQA model. This corresponds to the case where $\lambda=0$.
    \item SQuINT~\cite{selvaraju2020squinting}: Optimizes consistency by maximizing the similarity between the attention maps of pairs of questions. As such, it requires a VQA model that uses guided attention. 
    \item CP-VQA~\cite{tascon2022consistency}: Assumes entailment relations and uses a regularizer to improve consistency. 
\end{itemize}


\paragraph{VQA architectures:}
We show experiments using three VQA models depending on the dataset used. For experiments on Introspect, we make use of the BAN model~\cite{kim2018bilinear}, as its structure with guided attention allows the use of SQuINT. In addition, we evaluate the vision-language architecture LXMERT~\cite{tan2019lxmert} on this dataset to see how our approach can help improve state-of-the-art, transformer-based, VQA models too. For experiments on the DME dataset, we use the base model described in~\cite{tascon2022consistency}, which we denote by MVQA.

\subsection{Implementation details}
\label{subsec:imp_details}
\paragraph{LI-Model} We first pre-train BERT on SNLI for 5 epochs until it reaches a maximum accuracy of 84.32\% on that dataset. For this pre-training stage, we initialize BERT with the \textit{bert-base-uncased} weights and use a batch size of~16. We use a weight decay rate of 0.01 and the AdamW optimizer with learning rate $2\cdot10^{-5}$ without bias correction. The same setup was kept to finetune the model on a subset of 2'000 pairs of propositions from Introspect which were manually annotated (distribution of labels being: $\leftarrow 60\%, \leftrightarrow 17\%, - 12\%, \rightarrow  11\%$), and an additional 500 pairs were annotated for validation. Notice that LI-MOD is only necessary for the Introspect dataset, since for the DME dataset the implications annotations are available.

\paragraph{VQA models:}
For our base models, we use the official and publicly available implementations (BAN~\cite{jackroos2019}, LXMERT~\cite{tan2019lxmert} and MVQA~\cite{tascon2022consistency}) with default configurations. We re-implemented SQuINT~\cite{selvaraju2020squinting} and used the provided implementation of CP-VQA~\cite{tascon2022consistency}, reporting the best results which were obtained with $\lambda=0.1, \gamma=0.5$ for BAN and $\lambda=0.5, \gamma=1$ for MVQA. These parameters refer to the corresponding symbols of the original implementations. For SQuINT, we set the gain of the attention map similarity term to 0.5 for BAN and to 1.0 for MVQA. For Introspect, we train 5 models with different seeds for each parameter set and for DME we train 10 models with different seeds. To train LXMERT, BAN and MVQA, we use batch sizes of 32, 64 and 128, respectively. Regarding the VQA cross-entropy loss, we follow the original implementations and use soft scores for the answers in LXMERT, and categorical answers for BAN and MVQA.

%% file: sections/055_Results.tex
\section{Results}
\label{sec:results}

\begin{figure*}[!t]
\centering
\includegraphics[width=0.84\textwidth]{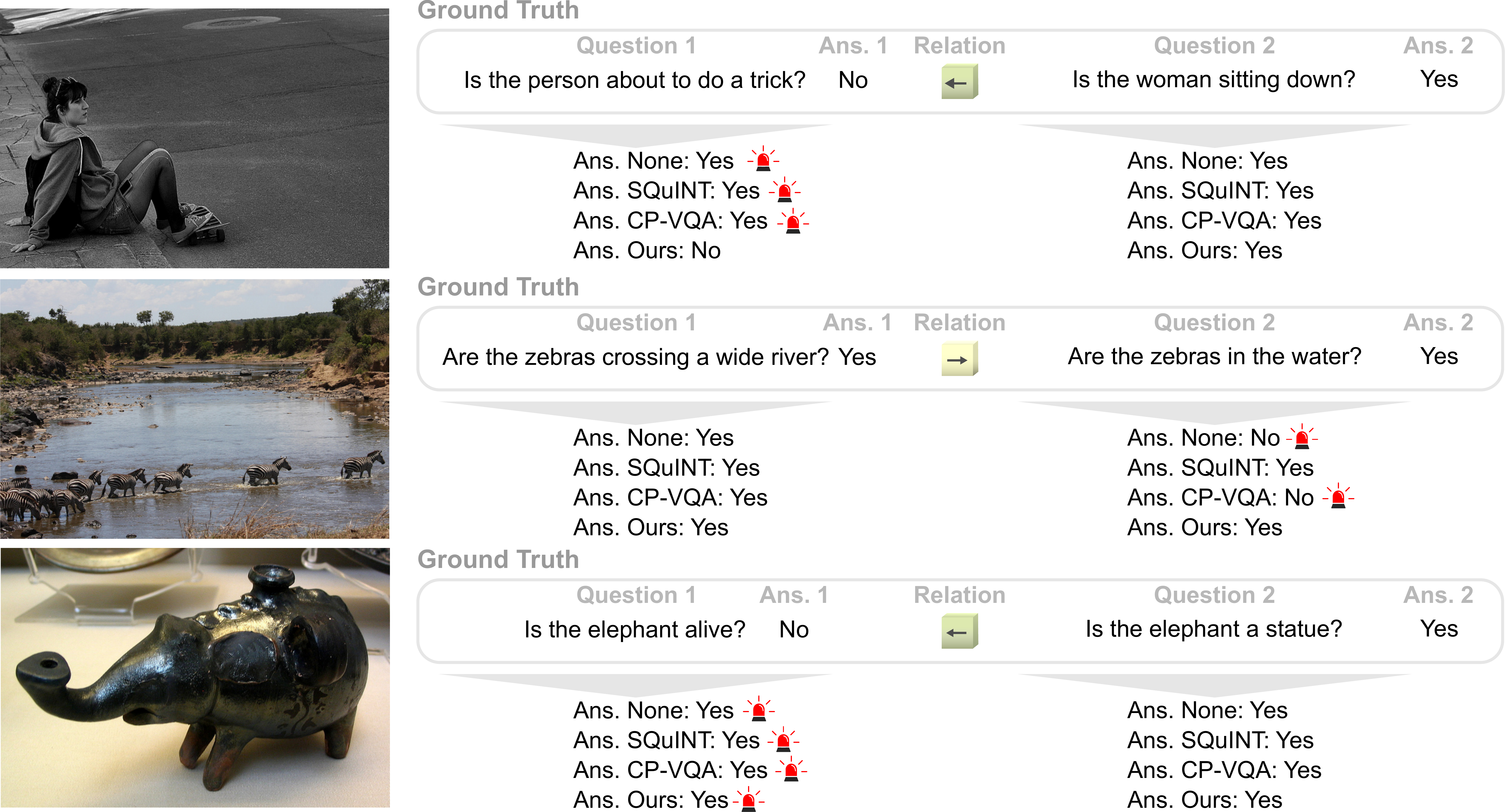}
\caption{Qualitative examples from the Introspect dataset using BAN as backbone. Red siren symbols indicate inconsistent cases.}
\label{fig:examples_introspect}
\end{figure*}
\begin{figure*}[!t]
\centering
\includegraphics[width=0.8\textwidth]{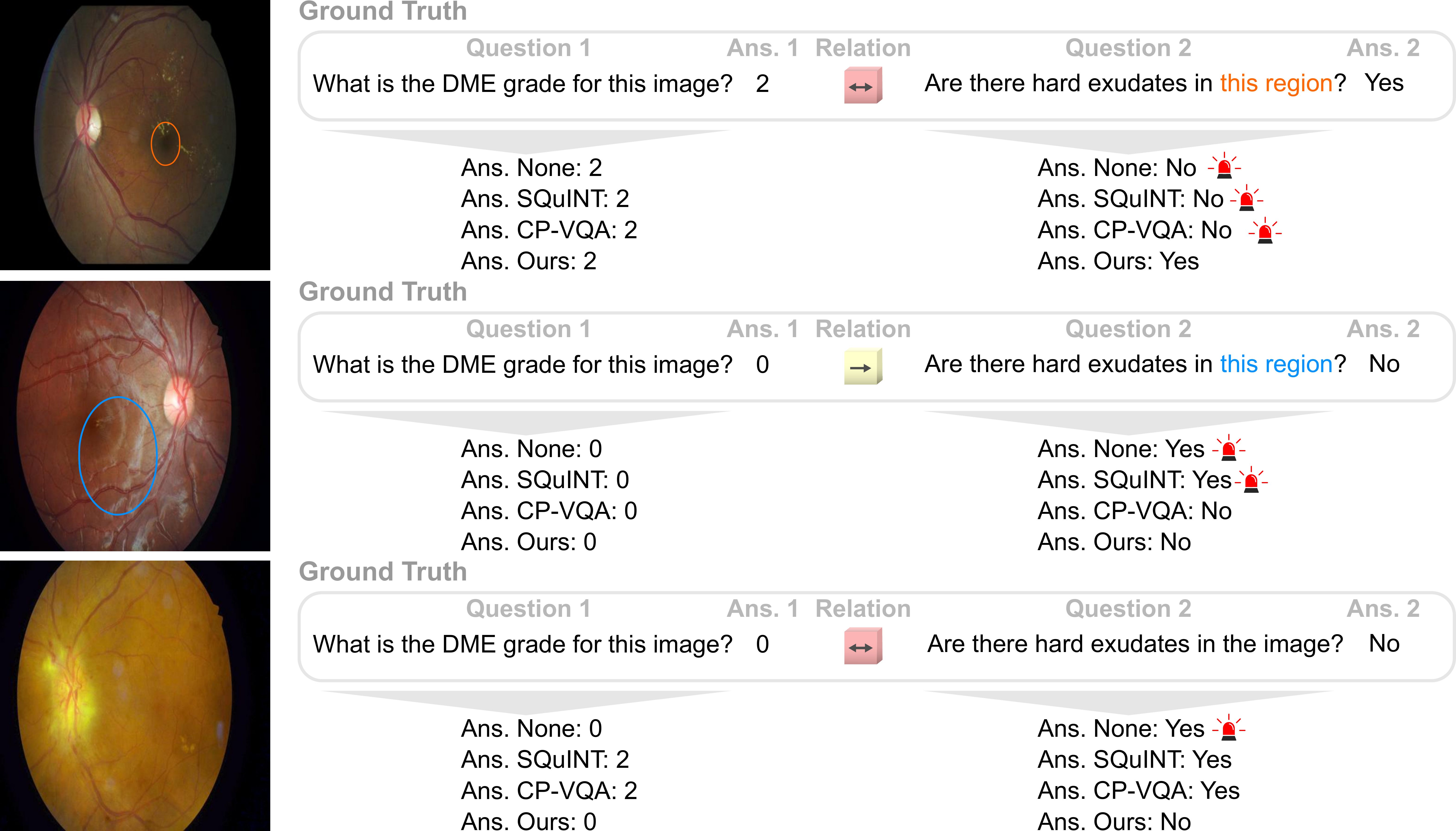}
\caption{Examples from the DME dataset and comparison of methods. Red siren symbols indicate inconsistent cases. DME is a disease that is staged into grades (0, 1 or 2), which depend on the number of visual pathological features of the retina. \textit{Top} and \textit{middle:} Although all methods correctly predict the answer to the first question, some inconsistencies appear when a necessary condition is false. \textit{Bottom}: Only the None baseline produces an inconsistency. Note that SQuINT and CP-VQA's answers do not produce inconsistent pairs because both questions were answered incorrectly, and those answers (``2" and ``yes") respect all known relations. 
}
\label{fig:examples_dme}
\end{figure*}

\paragraph{Performance comparison:}
For both datasets, we first compare the performance of our method against the baseline consistency methods in \cref{tab:results_introspect} and \cref{tab:results_dme}. In either case, we see that our method outperforms previous approaches, by not only increasing overall prediction accuracy but also by increasing consistency. In \cref{fig:examples_introspect} and \cref{fig:examples_dme}, we show illustrative examples of our approach on the Introspect and DME datasets, respectively (see additional examples in the Supplementary materials).


\begin{table}[!b ]
  \centering
  \begin{tabular}{@{}llcc@{}}
    \toprule
     Model & Cons. Method & Acc. & Cons. \\
     \midrule
     \multirow{4}{*}{BAN} & None & 67.14$\pm$0.10 & 69.45$\pm$0.17 \\
      & SQuINT~\cite{selvaraju2020squinting} & 67.27$\pm$0.19 & 69.87$\pm$0.45 \\
      & CP-VQA~\cite{tascon2022consistency} & 67.18$\pm$0.24 & 69.52$\pm$0.45\\
      & {\bf Ours} ($\lambda=0.01$) & {\bf 67.36$\pm$0.19} & {70.38$\pm$0.39} \\
    \midrule
    \multirow{5}{*}{LXMERT} & None & 75.10$\pm$0.10 & 76.24$\pm$0.63 \\
     & Random flip & 69.67$\pm$1.24 & 75.99$\pm$3.91\\
     & Flip first & 73.81$\pm$0.47 & 71.94$\pm$2.82\\
     & Flip second & 65.82$\pm$1.03 & {87.56$\pm$2.51}\\
     & {\bf Ours} & {\bf 75.17$\pm$0.08} & {78.75$\pm$0.21} \\
    \bottomrule
  \end{tabular}
  \caption{Results of different consistency methods on the Introspect dataset using two different VQA models: (top) BAN and (bottom) LXMERT. In the case of LXMERT, we show the impact of randomly flipping the answer of either the first or the second question for pairs detected as inconsistent using the relations from LI-MOD. Similarly, {\it flip first} and {\it flip second} refer to flipping the answer of the first and second question in inconsistent pairs, respectively.}
  \label{tab:results_introspect}
\end{table}

\begin{table*}[t]
\centering
\begin{tabular}{@{}llcccccr@{}}
\toprule
\multirow{2}{*}{Model} &  \multicolumn{1}{l}{\multirow{2}{*}{Consis. Method}}   & \multicolumn{5}{c}{Accuracy}  & \multicolumn{1}{c}{\multirow{2}{*}{Consistency}} \\ \cline{3-7}
\multicolumn{1}{c}{}         &\multicolumn{1}{c}{}                  & \multicolumn{1}{c}{all} & \multicolumn{1}{l}{grade} & \multicolumn{1}{c}{whole} & \multicolumn{1}{c}{macula} & \multicolumn{1}{c}{region} & \multicolumn{1}{l}{}                             \\ 
\midrule
\multirow{4}{*}{MVQA}     &   \multicolumn{1}{l}{None}          & \multicolumn{1}{c}{81.15$\pm$0.49}   & \multicolumn{1}{c}{78.17$\pm$2.07} & \multicolumn{1}{c}{83.44$\pm$1.87} & \multicolumn{1}{c}{87.25$\pm$1.20}  & \multicolumn{1}{c}{80.38$\pm$2.02}  & \multicolumn{1}{c}{89.95$\pm$3.20}                        \\ 

 &  \multicolumn{1}{l}{SQuINT~\cite{selvaraju2020squinting}}& \multicolumn{1}{c}{80.58$\pm$0.78}   & \multicolumn{1}{c}{77.48$\pm$0.40} & \multicolumn{1}{c}{82.82$\pm$0.74} & \multicolumn{1}{c}{85.34$\pm$0.87}  & \multicolumn{1}{c}{80.02}  & \multicolumn{1}{c}{89.39$\pm$2.12}               
\\ 

 &  \multicolumn{1}{l}{CP-VQA~\cite{tascon2022consistency}}& \multicolumn{1}{c}{83.49$\pm$0.99}   & \multicolumn{1}{c}{\textbf{80.69$\pm$1.30}} & \multicolumn{1}{c}{84.96$\pm$1.14} & \multicolumn{1}{c}{87.18$\pm$2.18} & \multicolumn{1}{c}{\textbf{83.16$\pm$1.09}}  & \multicolumn{1}{c}{94.20$\pm$2.15}  
\\ 

 &  
\multicolumn{1}{l}{{\bf Ours} ($\lambda=0.25$)}& \multicolumn{1}{c}{\textbf{83.59$\pm$0.69}}   & \multicolumn{1}{c}{80.15$\pm$0.95} & \multicolumn{1}{c}{\textbf{86.22$\pm$1.67}} & \multicolumn{1}{c}{\textbf{88.18$\pm$1.07}}  & \multicolumn{1}{c}{82.62$\pm$1.02}  & \multicolumn{1}{c}{{95.78$\pm$1.19}}  

\\ \bottomrule
                                                 
\end{tabular}
\caption{Comparison of methods on the DME dataset with common MVQA backbone. Accuracy and consistency is reported for all questions, as well as different medically relevant sub-question categories: grade, whole, macula and region.
}
\label{tab:results_dme}
\end{table*}

In \cref{tab:results_introspect} we also show the performance of the state-of-the-art LXMERT VQA model when combined with our proposed consistency method. In this case too, we see that our method provides increased performances via consistency improvements. Here we investigate the performance induced when flipping the answers of one of the members of each inconsistent pair at test time. Suppose implication labels are present, either by manual annotation or by LI-MOD. In that case, a trivial manner of correcting an inconsistent QA pair of binary answers is to flip or negate one of the answers. This is far simpler than our proposed method as it permits training the VQA model with the standard VQA loss. Having obtained the answers from the model when $\lambda=0$, we identify the inconsistent pairs using the relations predicted by our LI-MOD and then flip the answers (1) either randomly, (2) of the first QA or (3) of the second QA. By including the flipping baselines, we confirm that the added complexity in training our method results in improved accuracy compared to merely correcting inconsistencies post-hoc.To explain why the consistency can increase while the accuracy decreases, consider the following: An inconsistent QA pair guarantees that one of the two answers is incorrect, but correcting the inconsistency does not necessarily fix the incorrect answer. By flipping the correct answer, the inconsistency is corrected, thereby increasing the consistency but decreasing the accuracy. This phenomenon is particularly noticeable in the flipping baselines, as they fix inconsistencies without considering their correctness.


In general, we observe that training LXMERT with our consistency loss provides performance gains. Indeed, while random flipping based on LI-MOD clearly deteriorates the performance of LXMERT, so are flipping the first or second answers. This implies that our proposed method indeed leverages the predictions of LI-MOD to make LXMERT more consistent as it improves both model accuracy and consistency. 
\begin{figure*}[!t]
\centering
\includegraphics[width=0.85\textwidth]{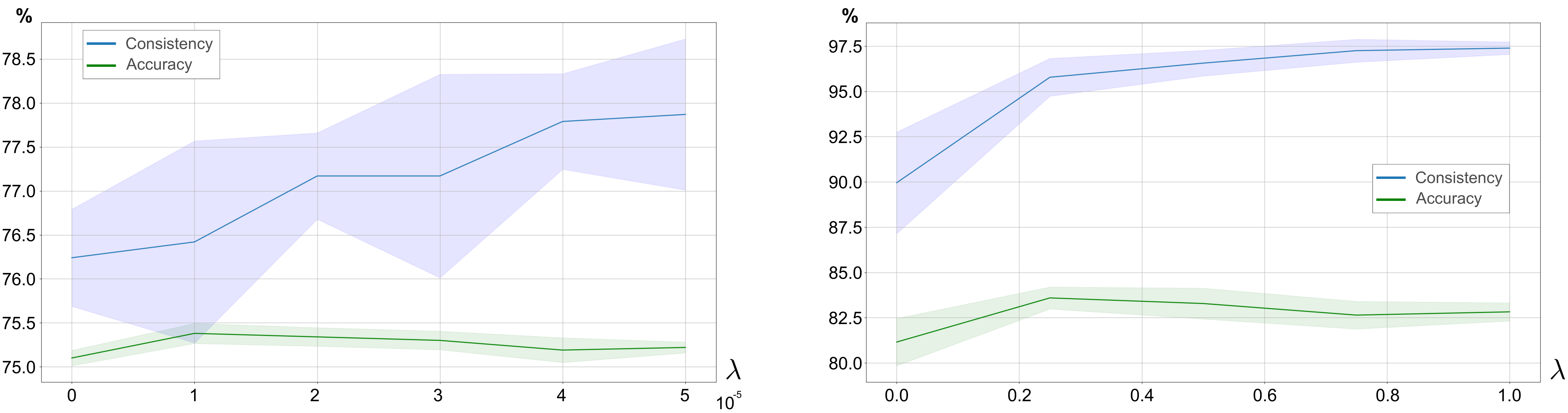}
\caption{Behavior of the accuracy and consistency as a function of $\lambda$ with 95\% confidence intervals. \textit{Left:} LXMERT trained on the Introspect dataset (5 models with random seeds for each value of $\lambda$). \textit{Right:} MVQA trained on the DME dataset (10 models with random seeds for each $\lambda$).
}
\label{fig:lambda}
\end{figure*}

\paragraph{Sensitivity of $\bm{\lambda}$:} We now show the sensitivity of our method and its relation to $\lambda$. We evaluate the performance of our method for different values of $\lambda$ to understand the behaviour of the performance, both in terms of accuracy and consistency. 

\cref{fig:lambda} shows the accuracy and consistency of LXMERT and MVQA for different values of $\lambda$. The difference in the ranges of the values is due to the relative magnitude of the loss function terms and depends on the used loss functions (\eg, binary and non-binary cross-entropy) and the ground-truth answer format (\ie, soft scores for LXMERT, as mentioned in \cref{subsec:imp_details}). 

In general, we observe a very similar behavior for accuracy, which increases and then slowly decreases as $\lambda$ increases. We sustain that the maximum value the accuracy can reach is established by the number of related pairs that are still inconsistent after training with $\lambda=0$. In other words, the limitations in size impose a limit for how much our method can improve the accuracy. For LXMERT on Introspect, for instance, our model corrected 4’553 (78.9\%) of the 5’771 existing inconsistencies and introduced new inconsistencies by mistakenly altering 1’562 (3.5\%) of the 44’111 consistent samples.

Regarding consistency, we observe a constant increase as $\lambda$ increases. The simultaneous decrease in accuracy as the $\lambda$ increases suggests that the relative weight of the consistency loss dominates so that the model no longer focuses on optimizing the cross-entropy. Since it is possible to be consistent without answering correctly, the optimization process results in an increase in consistency at the expense of accuracy for higher values of $\lambda$. However, it is clear from these results that there is a set of $\lambda$  values for which both accuracy and consistency improve.



\begin{figure}[!h]
    \centering
    \includegraphics[width=0.99\linewidth]{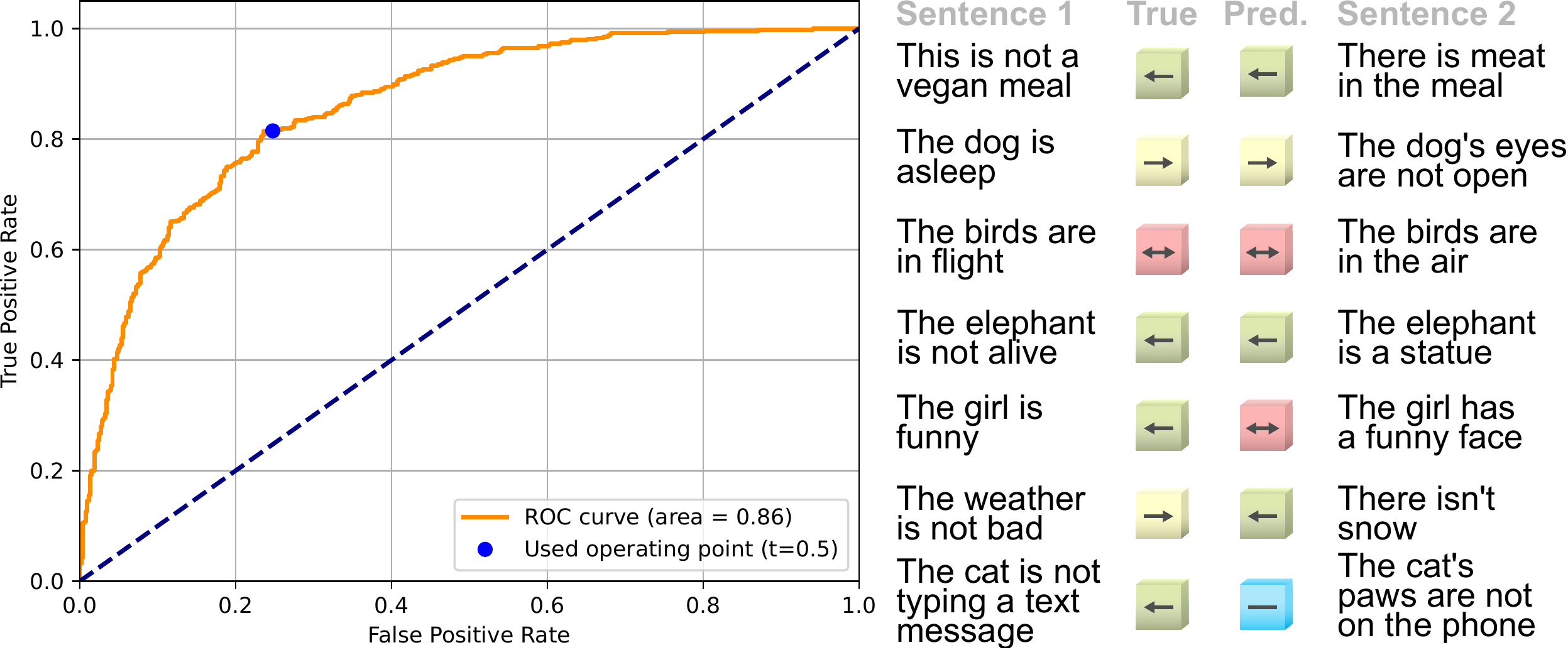}
    \caption{\textit{Left:} Receiver Operating Characteristic (ROC) for the entailment class of our LI-MOD in validation. \textit{Right:} Qualitative examples of LI-MOD's predictions.}
    \label{fig:roc_dot}
\end{figure}

\paragraph{LI-MOD performance:} We report that the finetuning of BERT on the subset of annotated relations from Introspect produced $78.67\%$ accuracy in the NLI task. We analyze the performance of this model for entailment and report an AUC value of 0.86, which indicates good generalization capability considering that only $\approx 2 \%$ of the dataset was annotated with relations. In addition, the amount of overlap in the QA pairs between the train and validation sets of the  Introspect dataset is only 1.12\% for binary questions. This shows that our LI-MOD is generalizing to variations in questions as well as to new combinations of QA pairs. \cref{fig:roc_dot} shows the ROC curve for entailment and examples of LI-MOD's predictions. Some of the observed sources of errors in LI-MOD include negations, unusual descriptions (\eg, a cat typing a text message), and image-specific references (\eg, ``is \textit{this} animal real?").

%% file: sections/06_Conclusion.tex
\section{Conclusion and future work}
\label{sec:conclusion}

In this paper, we propose a novel model-agnostic method to measure and improve consistency in VQA. We do so by integrating logical implications between pairs of questions directly in the training process. Additionally, we present a method to infer implications between QA pairs using a transformer-based natural language model. We conduct a series of experiments to verify the validity of our consistency loss in terms of generalizability and robustness against several baselines and across different datasets. Our results reveal the usefulness and applicability of our method to improve performances by reducing incoherence in responses. Future works include the creation of a larger dataset with human-annotated relations, which can then be used as general-purpose relations database for VQA training. 

%% file: sections/07_Acknowledgements.tex
\section*{Acknowledgements}
This work was partially funded by the Swiss National Science Foundation through grant 191983.